\documentclass[runningheads]{llncs}
\usepackage[T1]{fontenc}
\usepackage{graphicx}
\usepackage{multirow}
\usepackage{booktabs}
\usepackage{hyperref}
\usepackage{amsmath}
\usepackage{amssymb}
\usepackage{multirow}
\usepackage{color,soul}
\usepackage{subcaption}

\newcommand{\david}[1]{{\color{black}{#1}}}

\newcommand{\shuwen}[1]{{\color{black}{#1}}}
\newcommand{\lena}[1]{{\color{black}{#1}}}

\usepackage[misc]{ifsym}
    \makeatletter
\def\@fnsymbol#1{\ensuremath{\ifcase#1\or *\or $\Letter$\or \ddagger\or
   \mathsection\or \mathparagraph\or \|\or **\or \dagger\dagger
   \or \ddagger\ddagger \else\@ctrerr\fi}}
\newcommand{\printfnsymbol}[1]{%
  \textsuperscript{\@fnsymbol{#1}}%
}
    \makeatother
\begin{document}
%
\title{Detection of ADHD based on Eye Movements during Natural Viewing \footnote{This is a pre-print of an article to appear in Machine Learning and Knowledge Discovery in Databases. ECML PKDD 2022.}}
%
%

\author{Shuwen~Deng\inst{1}\thanks{Corresponding author: \email{shuwen.deng@uni-potsdam.de}} \and
Paul~Prasse\inst{1} \and
David~R.~Reich\inst{1} \and Sabine~Dziemian\inst{2} \and Maja~Stegenwallner-Sch{\"u}tz\inst{1,3} \and Daniel~Krakowczyk\inst{1} \and Silvia~Makowski\inst{1} \and Nicolas~Langer\inst{2} \and Tobias~Scheffer\inst{1}\and Lena~A.~J{\"a}ger\inst{1,4}}

\authorrunning{Deng \textit{et~al}.}
%
\institute{
    Department of Computer Science, University of Potsdam, Potsdam, Germany \and
    Department of Psychology, University of Zurich, Zurich, Switzerland \and
    Department of Inclusive Education, University of Potsdam, Potsdam, Germany \and
    Department of Computational Linguistics, University of Zurich,
    Zurich, Switzerland}

\toctitle{Detection of ADHD based on Eye Movements during Natural Viewing}
\tocauthor{Shuwen~Deng, Paul~Prasse, David~R.~Reich, Sabine~Dziemian, Maja~Stegenwallner-Sch{\"u}tz, Daniel~Krakowczyk, Silvia~Makowski, Nicolas~Langer, Tobias~Scheffer, Lena~A.~J{\"a}ger}
\maketitle              
\begin{abstract}
Attention-deficit/hyperactivity disorder (ADHD) is a neurodevelopmental disorder
that is highly prevalent and requires clinical specialists to diagnose. It is known that an individual's viewing behavior, reflected in their eye movements, is directly related to attentional mechanisms and higher-order cognitive processes. We therefore explore whether ADHD can be detected based on recorded eye movements together with information about the video stimulus in a free-viewing task.
To this end, we develop an end-to-end deep learning-based sequence model 
which we pre-train on a related task for which more data are available.
We find that the method is in fact able to detect ADHD and outperforms relevant baselines.
We investigate the relevance of the input features in an ablation study.
Interestingly, we find that the model's performance is closely related to the content of the video, which provides insights for future experimental designs.

\keywords{ADHD detection  \and eye movements \and free-viewing \and deep learning \and deep sequence models}
\end{abstract}

\section{Introduction}
\label{sec:introduction}

Attention-deficit/hyperactivity disorder (ADHD) is one of the most common neurodevelopmental disorders of childhood affecting approximately 5 to 13 percent of the children of an age cohort, depending on the diagnostic procedure used~\cite{Thomas2015,Willcutt2012,polanczyk2007worldwide}. ADHD is characterized by persistent inattention, high levels of hyperactivity, and impulsivity~\cite{AmericanPsychiatricAssociation2013}. 

The diagnosis of ADHD requires clinical assessment by specialists and typically involves self- and informant reports through clinical interviews and the use of rating scales.  Informant reports can be obtained from close family members, teachers, or partners, depending on the age of the candidate. Since the clinical assessment is heavily influenced by subjective reports and ratings, it also incurs the risk to reflect social or cognitive biases. The \textit{Strengths and Weaknesses of ADHD-Symptoms and Normal-Behavior (SWAN) rating scale}~\cite{swanson2012categorical} is a well-established screening tool based on a questionnaire that has to be filled out by parents or teachers. The SWAN scale registers symptoms of inattention, hyperactivity, and impulsivity yielding the so-called SWAN score. Specifically, the SWAN rating scale probes behaviors according to the full spectrum of symptom severity, which ranges from functionality to dysfunctionality~\cite{swanson2012categorical,brites2015development}.



The lack of comprehensive, objective assessment tools, developmental changes in the presentation of symptoms~\cite{Biederman2000}, and the high rates of co-morbidities~\cite{AmericanPsychiatricAssociation2013} present a major challenge to ADHD assessment and ultimately increases the risk of under- or overdiagnosis. While a false negative can lead to the denial of treatment, a false positive can lead to inappropriate treatment, both of which may have detrimental effects on an individual's ability to function at school, professionally and socially as well as on their overall well-being. 
This motivates the development of fully automatic screening tools that can be applied at large to people  at-risk or with a suspicion of having ADHD, thereby increasing the accessibility of ADHD screening opportunities as well as the objectivity of the screening method prior to specialist assessment. 

 Eye movements can be classified into so-called oculomotor events. These include fixations ($\approx$~200--300~ms), during which the eye is relatively still and visual information is obtained, and saccades, which are fast relocation movements of the eye gaze between any two fixations ($\approx$~30--80~ms)~\cite{holmqvist2011eye}. A sequence of fixations is referred to as a \textit{scanpath}. 
As eye movements are known to reflect cognitive processes including attentional mechanisms~\cite{justcarpenter1976,henderson2003human}, they are considered a \textit{window on mind and brain}~\cite{vanGompel2007eye}. For several decades, they have been used as a gold-standard measure in cognitive psychology~\cite{Rayner1998}. Researchers from the field of cognitive psychology typically treat eye movements as the dependent variable to investigate the effect of experimental manipulation of the stimulus and hence model it as the target variable. By contrast, more recent research has demonstrated the potential of treating eye movements as the independent variable (i.e., the model input) to infer the properties of the viewer. For example, it has been shown that eye-tracking data can be used to discriminate between different cognitive states \cite{henderson2013predicting}, personal traits \cite{hoppe2018eye}, or cognitive load \cite{Shojaeizadeh2019DetectingSystem}.  A major challenge in using eye movements to make inferences about a viewer is the high degree of individual variability in the eye-tracking signal. The dominance of individual characteristics in the eye-tracking data explains why machine-learning methods for viewer identification perform very well~\cite{lohr2020ijcb,makowski2021deepeyedentificationlive},
whereas models for other inference tasks typically perform at best at a proof-of-concept level or slightly above chance level. Another major challenge for the development of machine learning methods for the analysis of eye-tracking data is data scarcity. Since the collection of high-quality eye-tracking data is resource-intensive, only very few large data sets exist.


Differences in viewing behavior between individuals with and without ADHD have been found using eye-tracking tasks in which participants were required to make voluntary eye movements towards or away from a stimulus (so-called pro- or anti-saccade tasks)~\cite{munoz2004look,Klein2003}. 
These findings motivate our approach of developing a screening tool that processes each individual's eye movements and simultaneously takes into account information about the visual stimulus.

The contribution of this paper is fourfold. First, we provide a new state-of-the-art model to detect ADHD from eye movements in a natural free-viewing task and evaluate the performance of this model and relevant reference methods on a real-world data set. Second, we provide an extensive investigation of the relevance of the different input features in i) an ablation study and ii) by computing feature importances. Third, we demonstrate that transfer learning bears the potential to overcome the problem of data scarcity in eye-tracking research. Last but not least, we release a preprocessed free-viewing eye-tracking data set for the detection of ADHD. 

The remainder of this paper is structured as follows. Section~\ref{sec:related-work} discusses related work and Section~\ref{sec:problem_setting} lays out the problem setting. We develop a model architecture for the detection of ADHD in Section~\ref{sec:method}
and introduce the dataset in Section~\ref{sec:datasets}. In Section~\ref{sec:experiments} we present the experimental findings while in Section~\ref{sec:discussion}, we discuss the results. Section~\ref{sec:conclusion} concludes.

\section{Related Work}
\label{sec:related-work}
Machine learning methods have been applied for the purpose of ADHD detection to different types of diagnostic data; e.g., data of Conners’ Adult ADHD Rating Scales~\cite{christiansen2020use}, EEG signals~\cite{tor2021automated}, and functional Magnetic Resonance Imaging~(fMRI) data~\cite{deshpande2015fully} recorded in resting state.
The rapid development of affordable eye-tracking hardware offers new possibilities for non-invasive, rapid, and even implicit screenings that do not have to rely on self-, parent, or teacher reports. In the following section, we briefly review the work related to the use of machine learning methods with the purpose of identifying individuals with ADHD, with a particular focus on eye movement data.


ADHD detection has been conducted based on eye movements collected during different types of tasks, such as reading~\cite{de2019rule}, a 
reading span task~\cite{jayawardena2019eye},  or continuous performance tests~\cite{lev2022eye}. These tasks impose certain requirements on the participants in order to ensure the validity of the measurement; e.g., participants need to have already acquired a certain level of reading skills or have to understand and comply with complex task instructions. Moreover, it has been shown that under instructed conditions, eye movements are less affected by the type of content (e.g., emotional content) that is displayed than in natural viewing \cite{Kulke2022}. In order to reduce such limitations, first attempts have been made to detect ADHD on the basis of their eye movements in task-free viewing. In contrast to previous methods, this approach bears the potential to be applied already to very young children, which, in turn, allows them to gain access to treatment from a young age onwards. Early identification and treatment are crucial for mitigating the development of ADHD and its negative long-term consequences on individuals' functioning and overall well being~\cite{Jensen2001,Rubia2014}.

Galgani \textit{et al.}~\cite{galgani2009automatic} proposed three methods for ADHD detection through an image viewing task that they evaluated on participants with a  comparatively wide age range (9-59 years). Among these methods, the best-performing approach is based on the Levenshtein distance. This method uses regions of interest (ROI)-based alphabet encoding, which transforms a sequence of fixations into a sequence of symbols by assigning symbols to different ROIs. To classify a new instance, they compute the Levenshtein distance of the corresponding symbol sequence to 
instances in the ADHD group and the control group. A smaller average distance to a group indicates greater similarity to that group, and thus the corresponding group label is assigned to the instance. While this approach takes into account the spatial information of the sequence of fixations, it fails to consider the temporal information of fixations; i.e., the fixation duration. 

Instead of using a binary classifier for ADHD detection only, Tseng~\textit{et~al.}~\cite{tseng2013high} proposed a three-class classifier to differentiate between children with ADHD, children with fetal alcohol spectrum disorder, and control children, based on eye movements recorded during watching video clips of 15~minutes. They combined gaze features with visual saliency information of the stimulus computed with a saliency model.
However, 
they rely on engineered features that aggregate the eye gaze events over time (e.g., median saccade duration or saccade peak velocity) at the cost of the sequential information in the eye gaze signal not being used. 


More research has focused on using machine learning to detect other neurodevelopmental
disorders~\cite{jiang2017learning,wang2015atypical}. 
For example, Jiang~\textit{et~al.}~\cite{jiang2017learning} proposed to detect autism spectrum disorder (ASD) from eye-tracking data collected while viewing images, in which they used a neural network to explicitly model the differences in eye movement patterns between two groups. The main limitation of this method is that for each image only a fixed number of fixations are analyzed, which potentially causes information loss.

\section{Problem Setting}
\label{sec:problem_setting}
We study the problem of ADHD detection. While watching a video, the eye gaze of the $j$-th individual is recorded as a sequence of fixations, denoted as $P_{j} = \{(x_{1}, y_{1}, t_{1}), \ldots,  (x_{M}, y_{M}, t_{M})\}$, where $x_{m}$, $y_{m}$ are the $m$-th fixation location, $t_{m}$ is the fixation duration, and $M$ is the total number of recorded fixations. Provided a fixed video frame rate, we can use the temporal information to map the fixations to the corresponding video frames $V$, such that semantic information can be associated with eye-gaze. The training set consists of $\mathcal{D} = \{(P_1, V, c_1), \ldots, (P_J, V, c_J) \}$, where $P_j$ and $V$ represent the $j$-th individual's aligned fixation sequences and video frames, and $c_j$ is the label for whether an individual has ADHD. The objective is to train a classifier that identifies individuals with ADHD, which is a binary classification problem.


By varying the decision threshold for a learned model, we can plot the receiver operating characteristic (ROC) curve of the true positive rates versus false-positive rates, and finally compute the area under the curve (AUC) which is the area under the ROC curve and is used as a quantitative indicator of classification performance. We use the AUC as the evaluation metric, which is insensitive to the uneven distribution of classes.

\section{Method}
\label{sec:method}
In this section we introduce our model and the pre-training task used to initialize the weights for the final task of ADHD classification.

\subsection{Model}
\label{sec:model}
\begin{figure}[!t]
\begin{subfigure}[t]{0.99\textwidth}
\centering
\includegraphics[width=.85\textwidth,keepaspectratio]{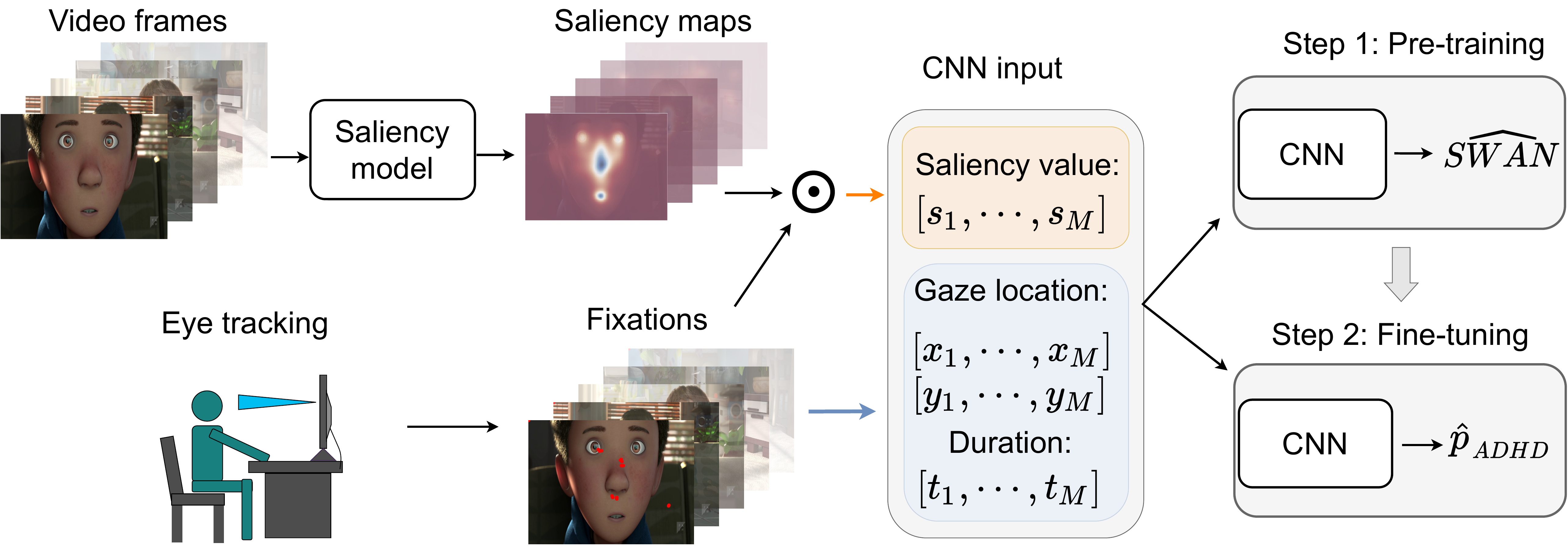}
\subcaption{Model overview.}
\vspace{8mm}
\end{subfigure}
\\
\begin{subfigure}[t]{0.99\textwidth}
\centering
\includegraphics[width=.5\textwidth,keepaspectratio]{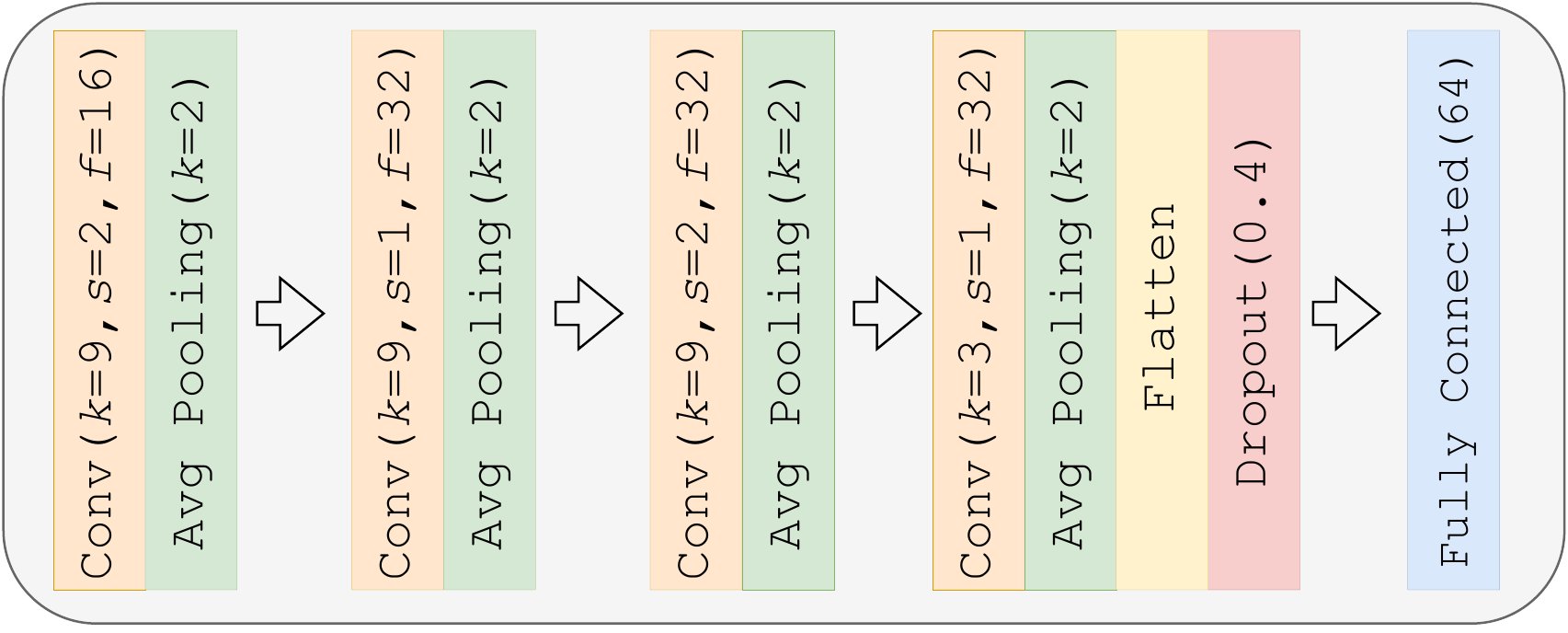}
\subcaption{Detailed view of the CNN.}
\end{subfigure}
\caption{Proposed network architecture. Panel (a) shows the complete architecture and Panel (b) shows the 1D-CNN denoted as ``CNN" in Panel (a). 
The model is pre-trained to predict the viewer's SWAN score (regression task) and fine-tuned for ADHD classification.}
\label{fig:method}
\end{figure}

We propose an end-to-end trained neural sequence model to classify gaze sequences as belonging to an individual with or without ADHD. Figure~\ref{fig:method} shows an overview of our proposed method. We preprocess the raw eye-tracking, which consists of 
horizontal and vertical screen coordinates recorded with a sampling rate of 60 or 120~Hz into sequences of fixations using the Dispersion-Threshold Identification algorithm \cite{salvucci2000identifying}. 
The model takes as input the eye gaze sequence (scanpath) and the video clip on which this scanpath has been generated. 

Based on our review of the literature, we hypothesized that the eye gaze of individuals with ADHD interacts differently with the visual stimulus in comparison to typically developing controls. 
We therefore use saliency maps to highlight possible regions of interest in a scene.
We use a state-of-the-art saliency model, DeepGaze~II~\cite{kummerer2017understanding}, to compute saliency maps for our video stimuli. DeepGaze~II uses VGG-19 features that were trained on an object recognition task \cite{simonyan2014very} and feeds them into a second network that is trained to predict a probability distribution of fixation locations on a given image. 

For each video frame $i$ of size $(W,H)$, the pre-trained DeepGaze~II model generates a saliency map  $S^{(i)} \in \mathbb{R}^{H\times W}$. We then apply min-max normalization to transform $S^{(i)}$ to the range of $[0,1]$. To extract the normalized saliency value of each fixation location, we create an extraction mask, $E_{m}^{(i)} \in \mathbb{R}^{H\times W}$, for the $m$-th fixation on the $i$-th video frame. More specifically, $E_{m}^{(i)}$ is generated by setting the fixation location to one and all other cells to zero. We then smooth the extraction mask with a Gaussian kernel (standard deviation~$\sigma$ = 1.5$^{\circ}$) and normalize it. The Gaussian kernel is applied to account for the parafoveal information intake around the center of the fixation \cite{holmqvist2011eye}. 
Eventually, the saliency value for the $m$-th fixation is given by:
\begin{equation}
    s_{m} = \textbf{1}_{H}\left(E_m^{(i)}\odot S^{(i)}\right)\textbf{1}_{W}^T,
\end{equation}
where $\odot$ is the Hadamard product and $\textbf{1}_d$ is an all-ones row vector of dimension~$d$.
In case a fixation spans multiple frames, we use the central frame for the saliency computation. \david{The extracted sequence of saliency values is concatenated with the fixation locations (represented in degrees of visual angle) and the fixation durations (see Figure~\ref{fig:method}).}
Finally, we apply z-score normalization to each of these feature channels.

\david{We then feed these feature channels}  into a 1D-convolutional neural network~(CNN) to perform the ADHD classification. \lena{Panel (b) of Figure~\ref{fig:method}} depicts the details of the CNN architecture. The CNN consists of four one-dimensional convolutional layers with rectified linear unit (ReLU) activation functions, followed by two linear fully-connected layers. We apply ReLU to the first layer and sigmoid to the last layer. Each convolutional layer is followed by a batch normalization layer and an average pooling layer with a pooling size of 2. 
The parameters $k$, $s$ \shuwen{and $f$}, 
specify the kernel size, the stride size, and the number of filters for the convolutions, respectively.
A dropout layer with a rate of 0.4 is added before the first dense layer to prevent over-fitting. 
Finally, the neural network is optimized using the binary cross-entropy metric.

\subsection{Pre-training}
\label{pretraining}
The number of data points from individuals with diagnosed ADHD and negatively-diagnosed controls in the dataset is limited. 
We therefore 
pre-train our model on a relevant task for which  more data is available. 
Specifically, we pre-train our neural network on a regression task predicting an individual's SWAN score. 
An individual's SWAN score is highly relevant to the diagnosis of ADHD; using the SWAN score to classify individuals with and without ADHD yields an AUC of 0.878 (standard error $ = 0.007$). We therefore capitalize on the SWAN score to enable the model to detect ADHD-related patterns in the eye movements and perform pre-training on the \textit{SWAN prediction dataset} (see Section~\ref{sec:datasets} for details on the datasets). 

For pre-training, we replace the sigmoid output unit with a linear output unit for the regression setting. We apply the mean squared error as loss function. The pre-trained weights are then used to initialize the ADHD classification model.

\section{Datasets}
\label{sec:datasets}
The data for this study is part of the ongoing Healthy Brain Network (HBN)\footnote{\url{https://healthybrainnetwork.org/}} initiative by the Child Mind Institute~\cite{alexander2017open}, establishing a biobank of multi-modal data of children and adolescents. The data analyzed here includes all participants of the HBN up to the 6th release. Participants from the 7th release were included if their data acquisition took place until the end of the season ``Spring 2019''.
\subsubsection{Naturalistic Stimuli Paradigm}
The tasks analyzed in this study include all free-viewing naturalistic stimuli paradigms of the test battery. 
Participants were shown four different age-appropriate videos with audio track: (1) an educational video clip (\textit{Fun with Fractals}, 2:43 min), (2) a short animated film (\textit{The Present}, 3:23 min), (3) a short clip of an animated film (\textit{Despicable Me}, 2:50 min), and (4) a trailer for a feature-length film (\textit{Diary of a Wimpy Kid}, 1:57 min). 
There were no instructions given for watching the videos. The order of the videos within the test battery was randomized for each participant except for \textit{The Present} always being shown last.

\subsubsection{Eye-Tracking}
Monocular eye gaze data of the right eye was recorded with an infrared video-based eye tracker (iView-X Red-m, SensoMotoric Instruments [SMI] GmbH, spatial resolution: 0.1$^\circ$, accuracy: 0.5$^\circ$). The eye gaze was recorded at a sampling rate of 60 Hz or 120 Hz, depending on the testing site. In between each task, the eye tracker was calibrated using a 5-point grid. 

\subsubsection{Participants}
The recruited participants were initially screened for having symptoms of any mental disorder. Clinical diagnoses were provided in accordance with the current edition of the Diagnostic and Statistical Manual of Mental Disorders (DSM-V) \cite{AmericanPsychiatricAssociation2013}, and based on a consensus by multiple licensed clinicians. 
A total of 1,246 participants were included in the study, whose tracker loss was less than 10\%. 232 participants (178 were male and 54 were female) with an age range of 6--21 years (mean age 9.97 years ± 3 years) were selected on the basis of having received an ADHD diagnosis (including the predominantly inattentive presentation, predominantly hyperactive-impulsive presentation, and combined presentation of ADHD) and having no past or current co-morbidity according to the DSM-V. These participants were assigned to the ADHD group. A group of 152 participants (71 were male and 81 were female) with an age range of 6--21 years (10.42 years ± 3.31 years) were assigned to the control group whose psychological assessment indicated no past or current presence of any mental disorder according to the DSM-V. All remaining 862 participants are included for hyperparameter tuning and pre-training the models. Hereafter, we refer to the subset of the data that contains recordings from the ADHD and control groups as \emph{ADHD classification dataset} and the subset  used for hyperparameter tuning and pre-training as \emph{SWAN prediction dataset}. Note that for some participants recordings are available only from a subset of the four videos, as detailed in Table~\ref{tab:demographic}. In addition to the diagnostic assessment, SWAN scores for participants were obtained through the SWAN scale as a measure of ADHD-related symptom severity~\cite{swanson2012categorical}.


\begin{table*}[]
    \caption{Number of individuals in the data. Numbers in parentheses show the number of ADHD (A) and healthy controls (C).}
    \label{tab:demographic}
    \begin{center}
        \begin{tabular}{l|l|l}
        \toprule
        Video           & ADHD classification dataset & SWAN prediction dataset\\ \hline

        Fun with Fractals & 67 (48 A, 19 C)     & 276\\
        The Present    & 159 (111 A, 48 C)    & 444\\
        Despicable Me  & 315 (187 A, 128 C)  & 656\\
        Diary of a Wimpy Kid & 340 (202 A, 138 C)  & 736\\
        \bottomrule
        \end{tabular}
    \end{center}
\end{table*}

\section{Experiments}
\label{sec:experiments}
In this section, we describe the experiments we conducted to evaluate our proposed approach and compare it with relevant reference methods. The code and data are available online.\footnote{\url{https://github.com/aeye-lab/ecml-ADHD}}

\subsection{Evaluation Protocol}
\label{sec:evaluation-protocol}

We perform 10 resamplings of 10-fold cross-validation while splitting the data by individuals. That means we test the model on the gaze sequence of unknown individuals, while the video stimulus has already been seen during training. We use the same data splits for all models to ensure a fair comparison. 

To evaluate the different models for different videos, we train a separate model for each video with and without pre-training. 
All neural network models are trained, using the Keras
and Tensorflow
libraries with the Adam optimizer
on an NVIDIA A100-SXM4-40GB GPU.

\subsection{Reference Methods}
\label{sec:baselines}
We compare our model with two relevant baseline methods. The first is the Levenshtein distance-based method proposed by Galgani~\textit{et~al.}~\cite{galgani2009automatic} (see Section~\ref{sec:related-work}).
This method was originally intended for the image domain,
which is not directly applicable to our video-based data. Therefore, instead of considering only the fixation sequences on a single image, we adapt it to a video-based classifier by calculating the Levenshtein distance based on the fixation sequences across the whole video.

Our second reference method is an approach proposed by Tseng~\textit{et~al.}~\cite{tseng2013high}, in which a support vector machine~(SVM) classifier is trained on aggregated engineered features extracted from eye gaze data collected while watching a 15-minute video composed of 2–4~s unrelated clip snippets. Tseng~\textit{et al.} focuses on ADHD detection in young children only, while the data collected from young adults is used as a reference to compute group-based features. Due to the lack of young individuals in the control group, we are not able to compute these group-based features and thus exclude them in our implementation.
Nevertheless, we implement all the remaining saliency- and gaze-based features and report the results of the model that is trained with recursive feature elimination as proposed by Tseng~\textit{et al.}~\cite{tseng2013high}.
To extract the saliency maps used by Tseng~\textit{et al.} we use the publicly available toolkit\footnote{\url{http://ilab.usc.edu/toolkit/}}.



\subsection{Hyperparameter Tuning}
\label{sec:hp-tuning}
To find the optimal parameter setup for the architecture introduced in Section~\ref{sec:model} we perform a random grid search using 5-fold cross-validation on the SWAN prediction dataset. 
Table~\ref{tab:hp} shows the search space for the parameters used during hyperparameter optimization where we restrict the kernel size of the convolutional layers to be less than or equal to the kernel size of the previous layer and the number of filters to be greater than or equal to the number of filters in the previous layer. Furthermore, the stride size is set to 1 when the kernel size is less than or equal to 5, and is restricted to be smaller than or equal to 2 when the kernel size is equal to 7. We use the data from the SWAN prediction dataset and predict the SWAN score to evaluate the stated hyperparameter configurations. The best performing configuration can be found in Figure~\ref{fig:method} and is used for all subsequent experiments.

\begin{table*}[]
\caption{Parameter used for hyperparameter optimization.}
\label{tab:hp}
\begin{center}
    \begin{tabular}{l|l}
    \toprule
    Parameter       & Search space            \\\hline
    \# conv layers   & \{3, 4, 5, 6, 7, 8, 9\} \\
    kernel size     & \{3, 5, 7, 9, 11\}      \\
    \# filters      & \{8, 16, 32, 64\}       \\
    stride size     & \{1, 2, 3\}             \\
    \# fully connected layers & \{1, 2, 3\}             \\
    \# hidden units  & \{8, 16, 32, 64\}       \\ 
    dropout rate    & \{0.2, 0.3, 0.4, 0.5, 0.6\}   \\
    pooling layer type & \{max pooling, average pooling\} \\
    \bottomrule
    \end{tabular}
\end{center}
\end{table*}

\subsection{Results}
\label{sec:experimental-results}

In Table~\ref{tab: model_results}, we present the evaluation results of our proposed models and the reference methods on all available videos. Except for the video \textit{Diary of a Wimpy Kid}, our proposed method (with and without pre-training) performs significantly above chance level ($p<0.05$). 
The best results are achieved for the video \textit{Fun with Fractals}. With regard to this video, the model trained from scratch achieves an AUC of around 0.58, and the pre-training further increases the performance by around 10\%.
Also for the \textit{Fun with Fractals} video, the results show that the proposed method with pre-training 
outperforms both  baselines. The comparison between the four videos further shows that all methods except for the model by 
Galgani~\textit{et~al.}~\cite{galgani2009automatic} perform best on the \textit{Fun with Fractals} video. This may indicate that certain properties of the stimulus video have an impact on how well the models can distinguish between individuals with ADHD and controls. 

\begin{figure*}[!ht]
	\centering
	  \includegraphics[width=0.8\textwidth,keepaspectratio]{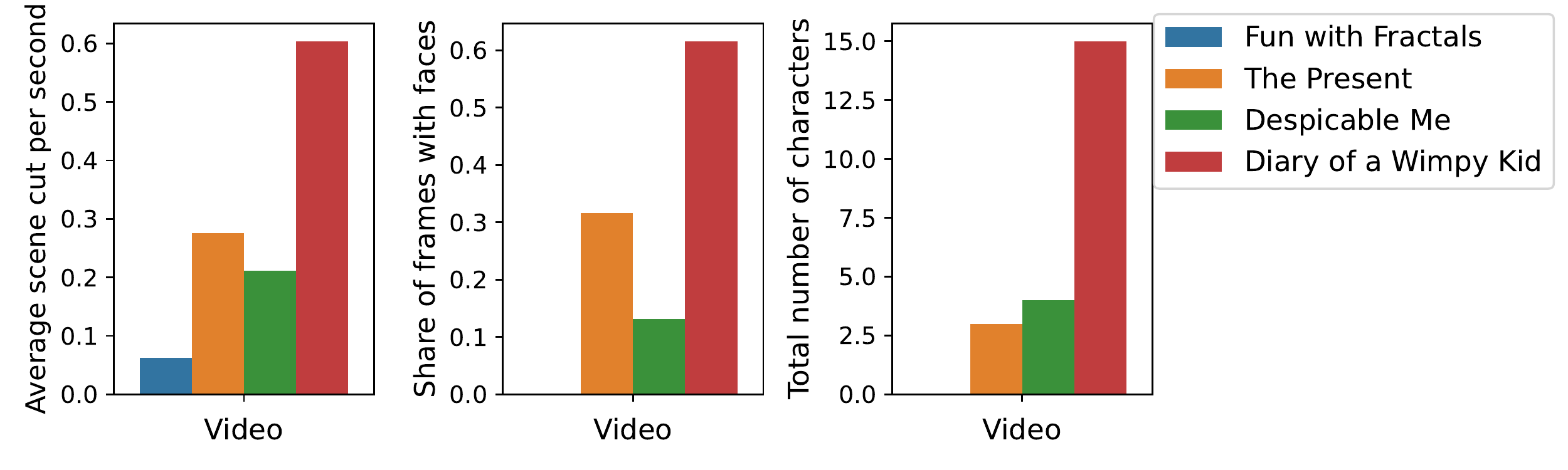}
	  \caption{Video Features of different video stimuli.}
	\label{fig:video_charac}
\end{figure*}

To characterize the differences between the videos, we extracted content-related features from each video which arguably quantify the video's degree of contingency~\cite{schwenzow2021understanding}: scene cut frequency, the proportion of frames showing at least one face, and the total number of characters that appear in the video (see Fig.~\ref{fig:video_charac}). The movie trailer \textit{Diary of a Wimpy Kid} has a large number of character appearances, a higher proportion of frames showing faces, and more frequent scene transitions. This arguably renders the video more engaging also for the ADHD group, which, in turn, may make their viewing behavior similar to the control group. The educational video \textit{Fun with Fractals}, by contrast, shows a low level of exciting content: The video mostly consists of relatively static scenes. According to the distinction between intact contingency-shaped and impaired predominantly self-regulatory processes of sustained attention among individuals with ADHD~\cite{Barkley1997}, their viewing behavior should be impacted by the video characteristics. Since the educational video contains less contingency, eye movements of individuals with ADHD may display more distinctive information for this video.

\begin{table*}[]
    \caption{AUC values $\pm$ standard error for ADHD detection of the CNN model with (CNN@Pre-tr.) and without (CNN@Scratch) pre-training. Galgani~\textit{et~al.} and Tseng~\textit{et~al.} refer to our re-implementation and adaptation to the data of their proposed method (see Section~\ref{sec:baselines}). The asterisk * indicates that the performance is significantly better than random guessing. The dagger $\dagger$ shows models significantly worse than the best model.}
    \label{tab: model_results}
    \begin{center}
        \begin{tabular}{l|l|l|l|l}
        \toprule
        Method & Fun with Fractals & The Present & Despicable Me & Diary of a  \\ 
        & & & & Wimpy Kid \\ \hline
        CNN@Scratch & 0.583 ± 0.026*  & 0.553 ± 0.017*      & 0.55 ± 0.01*     & 0.486 ± 0.01   \\
        CNN@Pre-tr. & \textbf{0.646 ± 0.025}*  & \textbf{0.554 ± 0.016}*     & 0.544 ± 0.01*    & 0.503 ± 0.01 \\
        Galgani~\textit{et~al.}~\cite{galgani2009automatic} & 0.33 ± 0.022$\dagger$  & 0.526 ± 0.017  & 0.523 ± 0.012*   & \textbf{0.515 ± 0.01}\\
        Tseng~\textit{et~al.}~\cite{tseng2013high} & 0.608 ± 0.023* & 0.418 ± 0.015$\dagger$            & \textbf{0.561 ± 0.011}*              & 0.465 ± 0.01      \\
        \bottomrule
        \end{tabular}
    \end{center}
\end{table*}

\subsection{Ablation Study and Feature Importance}
\label{sec:ablation}
In this section, we investigate the impact of each input feature of the proposed model with pre-training (CNN@Pre-tr) on the performance of the four different videos (see Table~\ref{tab:ablation}). In a second experiment, we 
look into the distribution of attribution scores using the attribution method DeepLIFT \cite{Shrikumar2017}, which is designed to explain model predictions.

Our proposed model consumes three different types of inputs:
saliency, fixation duration, and fixation location. Table~\ref{tab:ablation} shows the results for models trained without the saliency, fixation duration, and fixation location input in comparison to the model using all inputs. From Table~\ref{tab:ablation} we can conclude that removing any of the input channels lowers the model's performance for the \textit{Fun with Fractals} video. The drop in AUC is similar for each of the three components. Despite the drop in performance, we see that our model still outperforms both the model trained from scratch (CNN@Scratch) and the baseline models (see Table~\ref{tab: model_results}), which underlines the benefit of pre-training as well as the advantage of using multiple input channels. 
For the other three videos, we observe that removing one of the input channels does not have a systematic impact on the model's performance.  


\begin{table*}[]
    \caption{Results of the ablation study. The table shows AUC values $\pm$ standard error for our proposed model (CNN@Pre-tr.) trained with and without specific input features.}
    \label{tab:ablation}
    \begin{center}
        \begin{tabular}{l|l|l|l|l}
        \toprule  
        Model                & Fun with Fractals & The Present & Despicable Me                             &   Diary of a\\
        & & & & Wimpy Kid \\ \hline
        complete                  & \textbf{0.646 ± 0.025}* & 0.554 ± 0.016*   & 0.544 ± 0.01*   
                             & 0.503 ± 0.01\\
        w/o saliency          & 0.623 ± 0.026*          & \textbf{0.556 ± 0.016}*   & \textbf{0.545 ± 0.01}*                      & 0.494 ± 0.011\\
        w/o fix. duration          & 0.619 ± 0.027*          & 0.534 ± 0.016*                  & 0.536 ± 0.011*                              & 0.51 ± 0.01\\
        w/o fix. location         & 0.622 ± 0.026*          &  0.551 ± 0.015*                & 0.542 ± 0.01*                               &\textbf{0.526 ± 0.009}*\\
        \bottomrule
        \end{tabular}

    \end{center}
\end{table*}


  \begin{figure}[!ht]
    \subfloat[Individual with ADHD.\label{subfig:attribution-example-positive}]{%
      \includegraphics[trim=25 0 133 30,clip,width=0.425\textwidth,keepaspectratio]{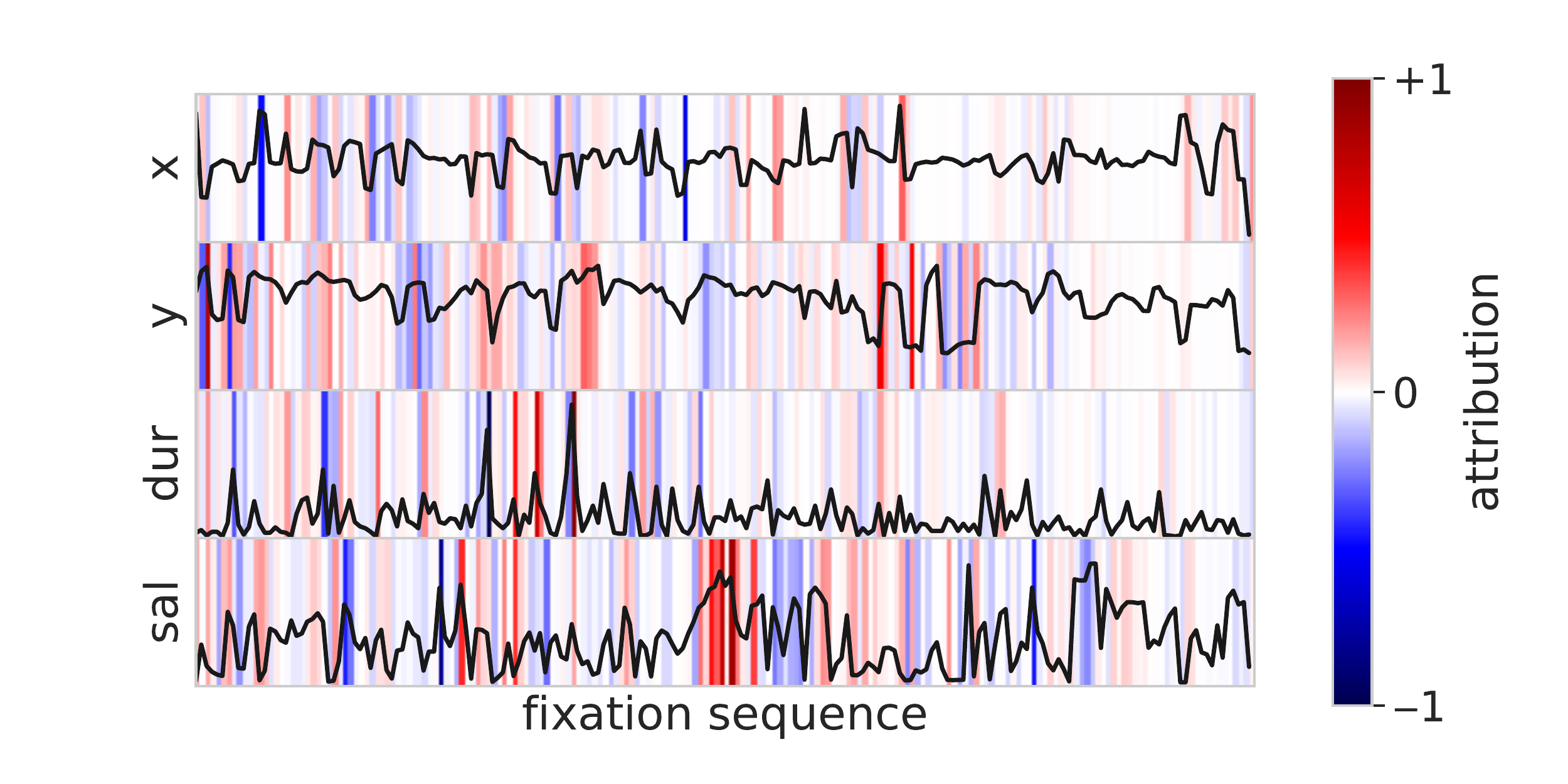}
    }
    \hfill
    \subfloat[Individual without ADHD.\label{subfig:attribution-example-negative}]{%
      \includegraphics[trim=25 0 133 30,clip,width=0.425\textwidth,keepaspectratio]{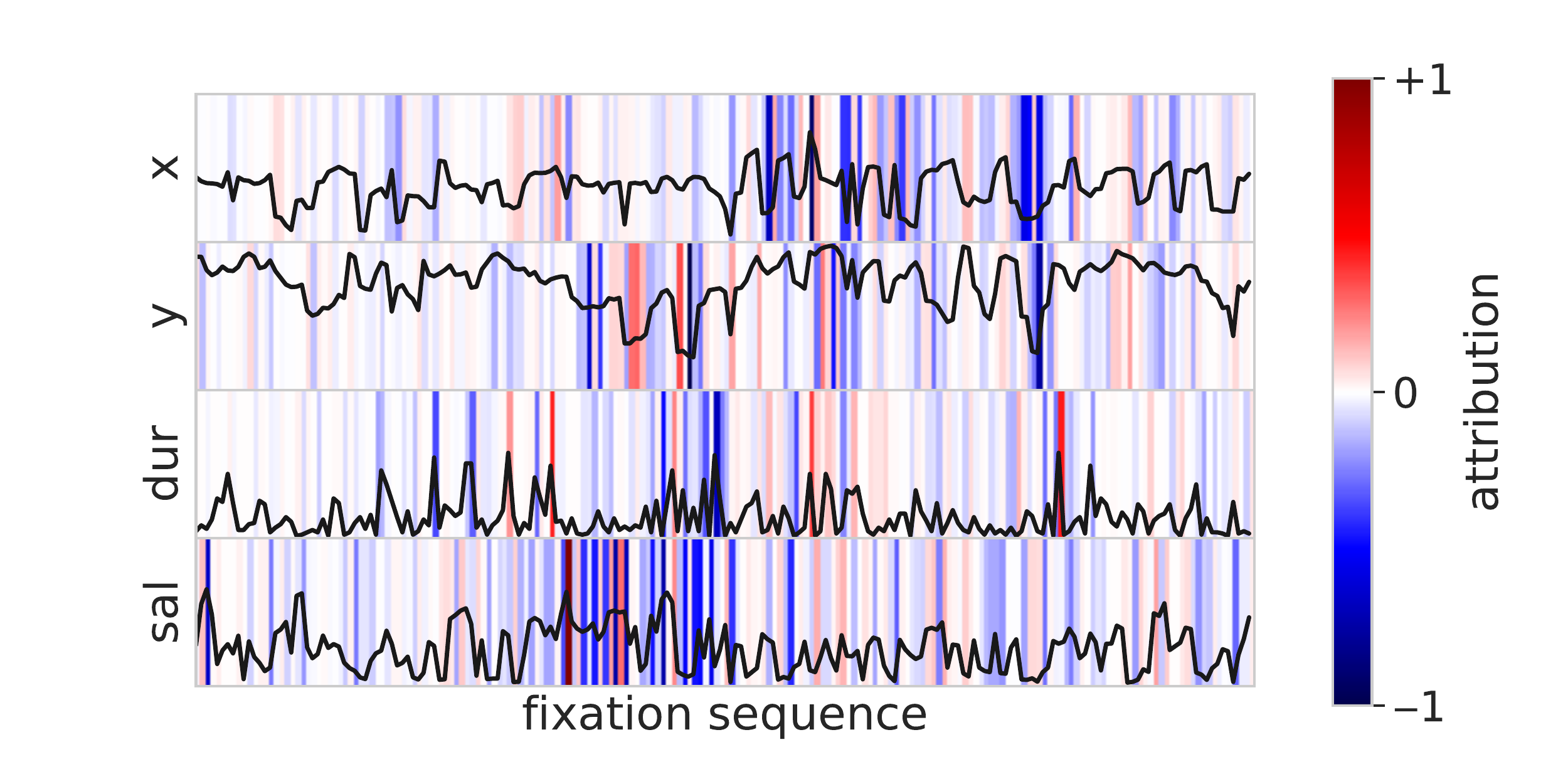}
    }
    \hfill
    \subfloat{
      \includegraphics[trim=600 0 0 30,clip,width=0.09\textwidth,keepaspectratio]{figures/attribution-example-2.pdf}
    }
    \caption{Example attributions for individual instances (ADHD and control group). The black lines represent the values of the input channel labelled on the y-axis. Red background colors show attribution relevance for ADHD, blue background colors show attribution relevance for the control group. Dark colors represent a high relevance, light colors a low relevance. }
    \label{fig:attribution-example}
  \end{figure}

In the second set of experiments, we investigate the feature importance of each input channel. To this end, we employ the post-hoc attribution method DeepLIFT~\cite{Shrikumar2017}, which belongs to the family of reference-based attribution methods. For each model prediction, these methods explain the difference in model output with respect to a previously chosen reference input. The explanations are provided as attribution values for each input feature and quantify the relevance to the model output. The resulting attributions can then be interpreted as a computationally less expensive approximation of SHAP values~\cite{Lundberg2016}.
Figure~\ref{fig:attribution-example} displays two example instances for an individual with and without ADHD, respectively.  These examples indicate that i) the attributions are spread relatively evenly over time and ii) the model uses all the available input channels. This is in line with our observation from the first part of the ablation study and confirms that all the input channels add valuable information.

\begin{figure*}[!ht]
	\centering
	  \includegraphics[width=0.75\textwidth,keepaspectratio]{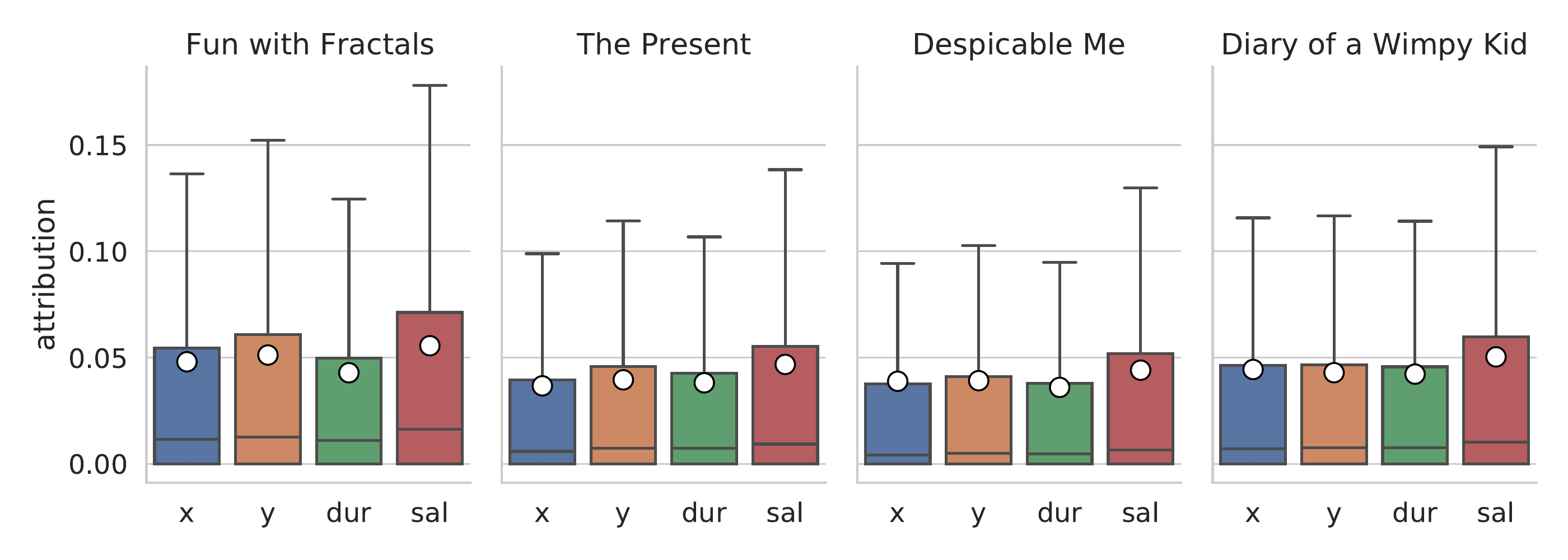}
	  \caption{Attribution box plot. Each video has a separate column with boxes for all four channels. Median value is represented by the horizontal black line in each box, mean value by the centered white dot. Whiskers are set to a 1.5 IQR value.}
	\label{fig:attribution-box-plot}
\end{figure*}

To determine the overall importance of each input channel we compute attributions for all instances of the ADHD classification dataset. We take the absolute values of the attributions and normalize the attributions for each instance to the range from zero to one. Figure~\ref{fig:attribution-box-plot} depicts the resulting box plot grouped by channels and videos. 
For all videos, the saliency channel is attributed with the highest relevance overall. While for the videos \textit{Despicable Me} and \textit{Diary of a Wimpy Kid} the fixation location channels  are about as relevant as fixation duration, the relevance is noticeably higher for the \textit{Fun with Fractals} video. Fixation duration is among the lowest attributed channels for all four videos. Note that the attribution of the two individual positional channels for the fixation location will add up to more relevance when not treated individually.

\section{Discussion}
\label{sec:discussion}
    
Our proposed model achieves state-of-the-art results in the detection of ADHD from eye movements. 
We developed a deep neural network that integrates a sequential stimulus, a video clip, with the corresponding gaze sequence. In contrast to previous research, we do not aggregate the eye gaze sequence over time, but rather developed a sequence model, that processes the unaggregated scanpath together with the saliency information of the visual stimulus that is currently around the center of the visual field (parafoveal vision). Our investigation of feature attributions revealed that the unaggregated information in the data is indeed used by the model. 
We have further demonstrated the advantage of pre-training the model on  a different task with additional data obtained from individuals diagnosed with other neurodevelopmental disorders. Whereas transfer learning approaches for \textit{predicting} eye movements exist~\cite{sood2021multimodal}, to the best of our knowledge, this is the first transfer learning approach processing eye-tracking data as input. As the recording of eye-tracking data is resource-intensive, data scarcity poses a major challenge to the development of machine learning methods for the analysis of eye movements. Our work demonstrates that transfer learning approaches with pre-training on a different domain or a different task offers the potential to fully exploit the information that is available in eye-tracking data.



%
The task-free nature of the viewing setting allows us to interpret eye movements to reflect differences in visual attention allocation between individuals with and without ADHD~\cite{Kulke2022}. With regard to clinical implications for ADHD-specific behavior, the model's successful prediction of ADHD group membership corroborates previous reports of distinctive eye movements displayed by individuals with ADHD in contrast to typically developing individuals.
This interpretation is also supported by comparisons between the different videos. When comparing the model's performance for the different videos, we noted a substantial improvement for the educational \textit{Fun with Fractals} video in comparison to the other three video clips. According to a distinction between intact contingency-shaped and impaired predominantly self-regulatory processes of sustained attention among individuals with ADHD~\cite{Barkley1997}, their viewing behavior should be impacted  by the video characteristics. Since the educational video contains less contingency,  eye movements from individuals with ADHD should exhibit a larger degree of dissimilarity from controls on this video. Our finding that differences in eye movements between individuals with and without ADHD are most pronounced on a less engaging video supports previous clinical findings~\cite{Silverstein2020}, according to which the demand of self-regulatory functioning impacts the performance of individuals with ADHD.

\section{Conclusion}
\label{sec:conclusion}

We developed a neural sequence model that reaches state-of-the-art performance in the classification of viewers with and without ADHD based on their eye gaze on a given video stimulus. 
Our method is widely applicable for the analysis of eye gaze data: It can be applied to any inference task that uses eye movements and a static or moving visual stimulus as input.  We have further demonstrated that the problem of data scarcity in eye-tracking research can be alleviated by pre-training on a different task for which more labeled data is available and by fine-tuning on the target setting. In conclusion, our method bears the prospective advantage of systematically exploiting eye movements in naturalistic settings for diagnostic purposes that includes, but is not limited to ADHD detection, and at the same time broadens our behavioral understanding of the disorder.


\subsubsection{Acknowledgements} This work was partially funded by the German Federal Ministry of Education and Research (grant 01$\vert$S20043) and a ZNZ PhD grant.
%
%
%
%
%

\end{document}